\documentclass[a4paper]{article}

\usepackage{makeidx}
\usepackage{graphicx}
\usepackage{amsmath,amssymb}
\usepackage{color}
\usepackage{hyperref}

\usepackage[english,USenglish]{babel}

\usepackage{natbib}
\setcitestyle{authoryear,open={(},close={)}}

\begin{document}

\title{RPBA -- Robust Parallel Bundle Adjustment Based on Covariance
  Information}

\author{Helmut Mayer\\Institute for Applied Computer Science\\ Bundeswehr University Munich\\
  Helmut.Mayer@unibw.de}

\date{}

\maketitle

\begin{abstract}
A core component of all Structure from Motion (SfM) approaches is
bundle adjustment. As the latter is a computational bottleneck for
larger blocks, parallel bundle adjustment has become an active area of
research. Particularly, consensus-based optimization methods have been
shown to be suitable for this task. We have extended them using
covariance information derived by the adjustment of individual
three-dimensional (3D) points, i.e., ``triangulation'' or
``intersection''. This does not only lead to a much better convergence
behavior, but also avoids fiddling with the penalty parameter of
standard consensus-based approaches. The corresponding novel approach
can also be seen as a variant of resection / intersection schemes,
where we adjust during intersection a number of sub-blocks directly
related to the number of threads available on a computer each
containing a fraction of the cameras of the block. We show that our
novel approach is suitable for robust parallel bundle adjustment and
demonstrate its capabilities in comparison to the basic
consensus-based approach as well as a state-of-the-art parallel
implementation of bundle adjustment. Code for our novel approach is
available on GitHub: \url{https://github.com/helmayer/RPBA}
\end{abstract}

\section{Introduction}
\label{sec:intro}

As ``Building Rome in a Day'' \citep{agarwal:09}, ``Building Rome on a
Cloudless Day'' \citep{frahm:10} and, particularly the more recent
``Reconstructing the World* in Six Days *(As Captured by the Yahoo 100
Million Image Dataset)'' \citep{heinly:15} show, Structure from Motion
(SfM) solutions can nowadays be obtained for huge numbers of
images. Yet, one has to note that these approaches deal with Community
Photo Collections \citep{goesele:10} which consist of many images
showing the same scenery. The goal is to link the images efficiently,
but usually only a very limited percentage of the images is linked and
the constructed separate blocks are much smaller than the photo
collections.

On the other hand, there are systems like ours \citep{mayer:12} which
in its recent form \citep{michelini:16,huang:19} can just deal with
about 6,000 images, but tries to link all of the given images even if
they have a small overlap and are strongly perspectively distorted due
to wide-baselines. The construction of the final block is done
hierarchically \citep{mayer:14}, similar to HyperSfM \citep{ni:12} and
the results are input to a system for fully three-dimensional (3D)
dense reconstruction for a large number of images \citep{kuhn:17}.

For both above scenarios a core part is the generation of large SfM
blocks. Besides the matching of images and the generation of
approximate values for (relative) pose estimation / orientation, it is
nowadays accepted that (bundle) adjustment of the block or sub-blocks
generated on the course to it is essential. For blocks up to a couple
of hundred cameras\footnote{We use the term ``cameras'' when
  discussing pose estimation / orientation of ``images''.} standard
solutions such as Sparse Bundle Adjustment (SBA) by
\citet{lourakis:09} are efficient enough.

Yet, for larger blocks the amount of memory required and the
computational complexity grow by the second and third power,
respectively. Thus, novel solutions are needed that make use of the
strength of modern computers, namely the parallel processing of a
larger number of threads.

Particularly, we present several extensions to the consensus-based
approach of \cite{eriksson:16}. There it has been shown how to compute
bundle adjustment in a distributed way. The basic idea is to split the
complete block into a number of sub-blocks related to the number of
available threads which are separately adjusted and the results are
synchronized by exchanging and averaging the 3D points.

\cite{ramamurthy:17}, \cite{zhang:17} and \cite{shen:18} have
demonstrated that one can additionally average the cameras or also
average the cameras instead of the 3D points. As there is much less
data for the cameras than for the 3D points, the latter has the big
advantage, that the amount of data to be exchanged is substantially
reduced. This is useful to keep the communication overhead at bounds
for really large data sets which are adjusted on computer clusters.

Our basic contribution consists in the insight, that the bundle
adjustment part of the fixed-point iterations in \citep{eriksson:16}
actually corresponds to the classical photogrammetric problem of
bundle adjustment taking into account 3D (ground) control
points. These are in our case the 3D points which link two or more
sub-blocks, named tie points (TPs) in this paper. Our insight leads to
two coupled improvements:

\begin{itemize}
  \item \cite{eriksson:16} weight the TPs globally by a penalty
    parameter $\rho$, for which the initialization is difficult and
    which also has to be changed over time to guarantee
    convergence. Opposed to this, we weight the TPs by the inverse of
    the full covariance matrix which is derived by adjustment
    (``intersection'') for each 3D point with the cameras fixed. By
    taking into account the covariance information of the TPs for
    those cameras, which are not in the respective sub-block, their
    influence on the adjustment is much better defined. By this means,
    we actually avoid the penalty parameter $\rho$ and still obtain a
    much better convergence than all the above approaches. For many
    blocks the much smaller number of iterations should make up for
    the larger amount of information we have to transfer between the
    bundle adjustment of the sub-blocks and the intersection of the 3D
    points, as we use points instead of cameras.
  \item While \cite{eriksson:16} just average the 3D points, we adjust
    the points with fixed cameras, which leads to improved coordinates
    and also to 3D covariance information. We also show that the
    adjustment of the points can be used for an effective and
    efficient robust estimation.
\end{itemize}

Additionally, as \cite{zhang:17} we have devised a meaningful way to
split a block into sub-blocks concerning the cameras based on
clustering the visibility graph, i.e., a graph where nodes correspond
to cameras and edges exist if two cameras share one or more 3D points.

The rest of the paper is organized as follows: In the next section we
give a short overview on bundle adjustment. This is followed by an
account of former work on parallel bundle adjustment in Section
\ref{sec:fw}. Section \ref{sec:cba} presents the basic ideas of
consensus-based bundle adjustment while Section \ref{sec:ecba}
introduces our extensions in detail, including robust
estimation. Extensive experiments characterizing the extensions and
making clear their advantages are given in Section \ref{sec:ep} before
ending up with the conclusion in Section \ref{sec:cc}.

\section{Bundle Adjustment}
\label{sec:ba}

This section gives an overview on bundle adjustment concerning issues
of interest for this paper. Early work on bundle adjustment dates back
more than sixty years by now \citep{brown:56,ackermann:62}. A report
on the state achieved forty years ago can be found in \citep{brown:76}
and \cite{triggs:99} give a comprehensive overview around the
millennium. Bundle adjustment related to the state of the art of
geometry and statistics can be found in the recent Manual of
Photogrammetry \citep{mcglone:13} and the textbook of
\cite{forstner:15}.

We follow in our notation \cite{jian:11}. Particularly, $\mathbf{P}_i$
with $i=1, \ldots, M$ denotes the camera parameters, $\mathbf{X}_j$,
with $j=1, \ldots, N$ the 3D points, and $\mathbf{x}_k$ with $k=1,
\ldots, K$ the 2D image coordinates for specific combinations of
camera $i$ and 3D point $j$, i.e., $k=(i,j)$.  The function
$\mathbf{g}_k(\mathbf{P}_i, \mathbf{X}_j)$ is used to project a 3D
point $\mathbf{X}_j$ by camera $\mathbf{P}_i$.
\begin{displaymath}
  \mathbf{v}_k = \mathbf{g}_k(\mathbf{P}_i, \mathbf{X}_j) - \mathbf{x}_k
\end{displaymath}
$\mathbf{v}_k$ is the residual between the projected 3D point and the
2D image point. Thus, the goal of bundle adjustment can be defined as
the minimization of the sum of the squared residuals (with $k=(i,j)$
indexing all existing combinations of cameras and points)
\begin{equation}
  f(\mathbf{P}, \mathbf{X}, \mathbf{x}) = \sum\limits_{k=1}^K [\mathbf{v}_k]^2 = \sum\limits_{k=1}^K [\mathbf{g}_k(\mathbf{P}_i,\mathbf{X}_j) - \mathbf{x}_k]^2 \;.
  \label{eq:v2}
\end{equation}

Because (\ref{eq:v2}) is nonlinear, it has to be linearized. This is done by
means of first-order Taylor expansion, assuming that appropriate
approximations for $\mathbf{P}_i$ and $\mathbf{X}_j$ are available:
\begin{equation}
  \sum_{k=1}^K [g_k(\mathbf{P}_i,\mathbf{X}_j) + 
  \frac{\partial \mathbf{g}_k(\mathbf{P}_i,\mathbf{X}_j)}{\partial \mathbf{P}_i} \mathrm d \mathbf{P}_i + 
  \frac{\partial \mathbf{g}_k(\mathbf{P}_i,\mathbf{X}_j)}{\partial \mathbf{X}_j} \mathrm d \mathbf{X}_j - \mathbf{x}_k]^2 \;.
  \label{eq:ta}
\end{equation}

To obtain a linear solution (\ref{eq:linear}), the derivatives in
(\ref{eq:ta}) are set to zero.  The resulting system consists of the
sparse design matrix $\mathbf A$ containing the Jacobian of the measurements
with respect to the cameras and 3D points, the vector
$\boldsymbol{\beta}$ concatenating the parameters of cameras and 3D
points, and the vector $\mathbf y$ consisting of the negative
measurement errors.
\begin{equation}
  \mathbf{A} \boldsymbol{\beta} = \mathbf{y} \;.
  \label{eq:linear}
\end{equation}

While (\ref{eq:linear}) can be solved directly, we employ the normal equations
(also sometimes called ``Hessian matrix'')
\begin{equation}
  \mathbf{N} \boldsymbol{\beta} = 
  (\mathbf{A}^{\mbox{\sf\scriptsize T}}\mathbf{A}) \boldsymbol{\beta} = 
  \mathbf{A}^{\mbox{\sf\scriptsize T}} \mathbf y
  \label{eq:normal}
\end{equation}
and compute $\boldsymbol{\beta} = \mathbf{N}^{-1} \mathbf{A}^{\mbox{\sf\scriptsize
    T}} \mathbf y$.

By this means one can introduce the estimated accuracy of the measured
image points in the form of a weight matrix.  Particularly, we employ
as weight the inverse of the approximate covariance matrix of the
measurements $\mathbf C$, leading to
\begin{equation}
  \mathbf{N} \boldsymbol{\beta} = 
  (\mathbf{A}^{\mbox{\sf\scriptsize T}}\mathbf{C}^{-1}\mathbf{A}) \boldsymbol{\beta} = 
  \mathbf{A}^{\mbox{\sf\scriptsize T}} \mathbf{C}^{-1} \mathbf y \; .
  \label{eq:normalw}
\end{equation}

$\mathbf C$ is a positive definite block diagonal matrix consisting of
$2 \times 2$ blocks describing the variance of the measured points in
$x$- and $y$-direction as well as their $x$-$y$ covariance.

As already in \citep{brown:76}, we split up the design matrix in a
part for 3D points $\mathbf{A}_X$ and a part for the cameras
$\mathbf{A}_C$. This results in the following (symmetric) matrix
$\mathbf{N}$ and its inverse $\mathbf{M}$
\begin{displaymath}
  \mathbf{N} =
  \left[
    \begin{array}{cc}
      \mathbf{N}_{XX} & \mathbf{N}_{XC} \\
      \mathbf{N}_{XC}^{\mbox{\sf\scriptsize T}} & \mathbf{N}_{CC}
    \end{array}
  \right] \hspace{0.3cm} \mbox{and} \hspace{0.3cm}
  \mathbf{M} = \mathbf{N}^{-1} = 
  \left[
    \begin{array}{cc}
      \mathbf{M}_{XX} & \mathbf{M}_{XC} \\
      \mathbf{M}_{XC}^{\mbox{\sf\scriptsize T}} & \mathbf{M}_{CC}
    \end{array}
  \right] \; .
\end{displaymath}

We employ the Schur complement to obtain the Reduced Camera System --
RCS \citep{jeong:12} $\mathbf{S} = \mathbf{N}_{CC} -
\mathbf{N}_{XC}^{\mbox{\sf\scriptsize T}} \mathbf{N}^{-1}_{XX}
\mathbf{N}_{XC}$. The solution of $\mathbf{S} \boldsymbol{\beta}_C =
\mathbf{y}_C$ (with ${y}_C$ the measurement errors reduced to the
cameras) concerning the cameras is at the core of conventional bundle
adjustment. The computation of $\mathbf{M}_{XX} = \mathbf{N}_{XX}^{-1}
+ \mathbf{N}_{XX}^{-1} \mathbf{N}_{XC} \mathbf{M}_{CC}
\mathbf{N}_{XC}^{\mbox{\sf\scriptsize T}} \mathbf{N}_{XX}^{-1}$ can be
done very efficiently, as it only involves the inversion of $3 \times
3$ matrices in the block diagonal matrix $\mathbf{N}_{XX}$ and
multiplications with $3 \times 6$ and $6 \times 6$ matrices.

For optimizing the solution, often the Levenberg-Marquardt -- LM
\citep{levenberg:44} algorithm, i.e., a damped Newton approach, is
used.  In its basic form the normal equations are augmented $\mathbf{N}_{aug} =
\mathbf{N} + \lambda \mathbf{I}$ by a damping factor $\lambda$ added to the diagonal
elements of the normal equations ($\mathbf{I}$ is the unit matrix).

While $\mathbf{M}_{CC}$ is the covariance of the cameras, its
computation by means of matrix inversion entails a large effort, when
the matrix becomes larger. Thus, in nearly all cases instead of
computing the inverse, the equation system is solved directly, e.g.,
by means of Cholesky factorization. For this it is mostly also
taken into account that the matrix is more or less sparse.

\section{Former Work}
\label{sec:fw}

This section is concerned with general former work on parallel bundle
adjustment. Specific work using consensus-based approaches such as
\citep{eriksson:16,zhang:17,ramamurthy:17} are treated in the next
section.  The subsection also introduces recent developments to
improve the speed and stability of bundle adjustment.

Work on parallel adjustment goes back to \cite{helmert:80}, though
\cite{agarwal:14} note that the proposed blocking only works well for
planar graphs, which is often not the case for SfM. In the recent two
decades, work on efficient bundle adjustment has first focused on
splitting the block into sub-blocks for reducing the memory
consumption and on efficient implementation. Concerning the former,
\cite{ni:07} have dealt with out-of-core bundle adjustment. Their
basic idea is to split the whole block into sub-blocks which are
solved independently and then re-aligned. This reduces memory
consumption as the size of the RCS is increasing quadratically in
relation to the number of cameras. Yet, the way chosen for
re-alignment in \citep{ni:07} is only suitable for blocks which can be
split into stable rigid sub-blocks whose relation can be approximated
well by a 3D similarity transformation. Opposed to this, our novel
approach is much more flexible, because we use TPs for transformation
and, thus, the sub-blocks can be deformed in a much more complex and
flexible way.

The first publicly available efficient implementation for bundle
adjustment is arguably Sparse Bundle Adjustment (SBA) by
\cite{lourakis:09} employing LM together with Cholesky
factorization. SBA is a very good baseline implementation, but because
it only makes use of the sparsity of the Jacobian and does not deal
with the sparsity of the normal equations, it is not suitable for
really large blocks.

\citep{agarwal:10} employ so-called inexact Newton solvers, which
basically means that Conjugate Gradient (CG) solvers are used. Yet, CG
is only efficient when combined with a suitable preconditioner.  Just
inverting the elements on the main diagonal does not entail much
effort, but the preconditioning is weak. Thus, experience has shown
that inverting the blocks for the cameras on the main diagonal, the
so-called block-Jacobi preconditioner, gives a good trade-off between
the complexity of the computation of the inverse blocks and the
effectiveness of preconditioning.

An implementation devised by the first author of \citep{agarwal:10}
originally for solving non-linear least-squares problems but with a
focus on bundle adjustment is the Ceres Solver \citep{ceres-solver}.
It contains also sparse Schur complement-based solvers specifically
designed for bundle adjustment.

The good trade-off of the block-Jacobi preconditioner is confirmed by
\cite{jeong:12} which add as argument in favor of the block-Jacobi
preconditioner which they use to solve the RCS also the ease of
implementation. We took particularly the latter as argument to also
use the block-Jacobi preconditioner, even though we note that
\cite{jian:11} have shown that more complex subgraph preconditioners
can give better results.

Additionally, \cite{jeong:12} deal with the sparsity in the RCS by
re-ordering the cameras in a way which minimizes the fill-in during
factorization. The latter had been already addressed in
\citep{snay:76,ayeni:80,kruck:83}. Also \cite{carlone:14} treat the
sparsity by clustering points visible in multiple cameras for a more
optimal representation.  While both means are useful, we do not employ
them, because we work with (much) smaller sub-blocks as introduced in
the next two sections which strongly reduces the problems with
sparsity in the sub-blocks.

\citet{hansch:16} focus on the implementation on massively parallel
Graphics Processing Units (GPUs) comparing (individual adjustment of
each camera) resection / intersection (triangulation) schemes with
non-linear CG and LM, concluding that all methods have their pros and
cons. The work of \citep{balasalle:18} is concerned with large scale
parallel bundle adjustment but with a focus on the specific problems
of satellite images and particularly planetary imaging.

Further recent work on bundle adjustment comprises \citep{zhu:14} on
the local readjustment of the camera poses and \citep{zhu:18} as an
example for distributed motion averaging. Finally, \cite{fusiello:15}
introduce bundle block adjustment by Generalized Anisotropic
Procrustes Analysis with advantages concerning convergence from
far-off approximate values. Unfortunately, only results for small data
sets are given and it is unclear how the approach scales, thus, we did
not consider it for our approach.

\section{Consensus-Based Bundle Adjustment}
\label{sec:cba}

Our approach for parallel bundle adjustment is based on consensus. In
the seminal paper of \cite{eriksson:16} consensus is achieved by
proximal splitting. Here we follow \cite{zhang:17} and
\cite{ramamurthy:17} and employ ADMM -- Alternating Direction Method
of Multipliers \citep{bertsekas:89}. The problem is split into a
number of sub-problems. The global consensus problem to be solved is
according to the equations from \citep{zhang:17}:

\parbox{10cm}
{\begin{eqnarray*}
  \mbox{minimize} & \sum\limits_{i=1}^n f_i(\mathbf x_i) \\
  \mbox{subject to} & \mathbf x_i = \mathbf z, i = 1, \ldots, n \;.
\end{eqnarray*}} \hfill
\parbox{1cm}
{\begin{eqnarray}
    \label{eq:admm0}
\end{eqnarray}}

The idea is to split the problem defined by function $f$ into $n$
parts which can be solved in parallel in $n$ threads, but are linked
by constraining the solutions to be consistent by referring to the
global variable $\mathbf z$. Usually $n$ is closely related to the
number of threads available on a computer.

The iterative solution with the ADMM algorithm according to the final
version in \citep{zhang:17} starts for time/iteration number $t + 1$
with optimizing a combination of the function in the individual parts
$f_i(\mathbf x_i^t)$ together with a term representing the distance
between the parameters $\mathbf x_i$ for the individual parts and
their global estimates for time $t$. The latter consist of the average
$\mathbf z_i$ of the $\mathbf x_i$ computed in the second step as well
as a difference weighted by the penalty parameter $\rho$ driving the
solution in the third step

\begin{eqnarray}
  \mathbf x_i^{t+1} & = & \arg \min_{\mathbf x_i} \Bigl( f_i(\mathbf x_i^t) +
  \frac{\rho}{2}  \| \mathbf x_i^t - (\mathbf z^t - \mathbf u_i^t) \|_2^2 \Bigr) \label{eq:admm4} \\
  \mathbf z^{t+1} & = & \frac{1}{n} \sum\limits_{i=1}^n \mathbf x_i^{t+1}   \label{eq:admm2} \\
  \mathbf u_i^{t+1} & = & \mathbf u_i^t + \mathbf x_i^{t+1} - \mathbf z^{t+1} \;.
  \label{eq:admm5}
\end{eqnarray}

The above framework can be used to split the bundle adjustment
(\ref{eq:v2}) into $L$ sub-blocks with the goal ($P_i^l$ and $X_j^l$
refer to the camera parameters and 3D points contained in sub-block
$l$, respectively)

\parbox{10cm}
{\begin{eqnarray*}
  \mbox{minimize} & \sum\limits_{l=1}^L f_l(\mathbf{P}^l,\mathbf{X}^l,\mathbf{x}) \\
  \mbox{subject to} & \mathbf{P}_i^l = \mathbf{P}_i, i = 1, \ldots, M, l = 1, \ldots, L \\
                    & \mathbf{X}_j^l = \mathbf{X}_j, j = 1, \ldots, N, l = 1, \ldots, L \; .
\end{eqnarray*}} \hfill
\parbox{1cm}
{\begin{eqnarray}
    \label{eq:admm6}
\end{eqnarray}}

\begin{figure}
   \begin{center}
      \leavevmode \unitlength 1cm 
      \includegraphics[width=12cm]{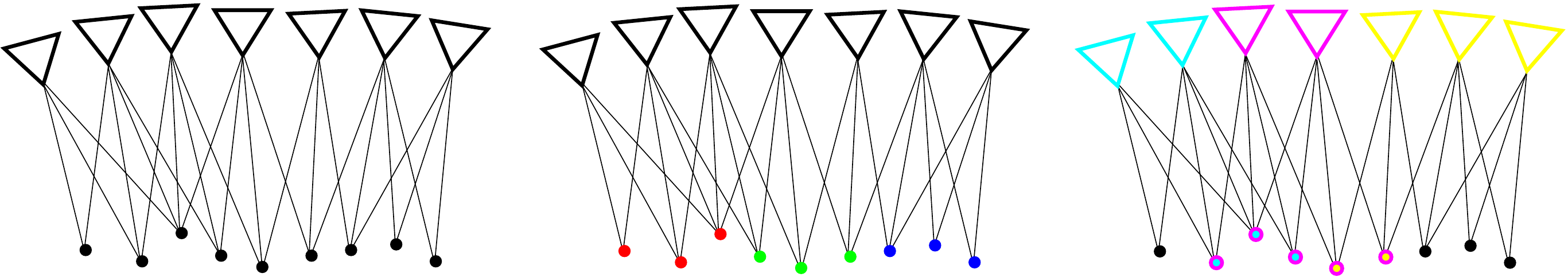}
   \end{center}
   \caption[]{A block (left) is split into sub-blocks with unique
     points according to \cite{zhang:17} (center) and with unambiguous
     cameras -- our case (right). The colored points on the right are
     the tie points contained in different sub-blocks.}
   \label{fig:sub-blocks}
\end{figure}

While in \cite{zhang:17} each 3D point is just contained in one
sub-block without overlap, we split the cameras unambiguously (Figure
\ref{fig:sub-blocks}). I.e., for the remainder we assume that every
camera is just contained in one sub-block and the relation between
sub-block $l$ and camera $\mathbf{P}_i$ with index $i$ is unique. On
the other hand, 3D points can be visible in more than one
sub-block. We term these points tie points (TPs).  $\mathbf{X}_j^l$
refers to a 3D point in a sub-block and for a TP $\mathbf{X}_j$
represents the average over all sub-blocks while
$\tilde{\mathbf{X}}_j^l$ drives the ADMM iteration like $u_i$ in
(\ref{eq:admm5}).

\begin{eqnarray}
 (\mathbf{P})^{t+1}, & & \nonumber \\
(\mathbf{X}^l)^{t+1} & = & \arg \min_{\mathbf{P}, \mathbf{X}^l} \Bigl( f_l((\mathbf{P})^t,(\mathbf{X}^l)^t,\mathbf{x})
  + \frac{\rho}{2}  \| (\mathbf{X}^l)^t - (\mathbf{X})^t + (\tilde{\mathbf{X}}^l)^t \|_2^2 \Bigr) \label{eq:admm7} \\
  (\mathbf{X}_j)^{t+1} & = & \frac{1}{L} \sum\limits_{l=1}^L (\mathbf{X}_j^l)^{t+1}   \label{eq:admm8} \\
  (\tilde{\mathbf{X}}_j^l)^{t+1} & = & (\tilde{\mathbf{X}}_j^l)^t + (\mathbf{X}_j^l)^{t+1} - (\mathbf{X}_j)^{t+1} \;.
  \label{eq:admm9}
\end{eqnarray}

In (\ref{eq:admm7}) cameras as well as points are jointly optimized,
but separately for each sub-block. (\ref{eq:admm8}) averages the
coordinates of the TPs and, finally, in (\ref{eq:admm9}) an
improvement is computed for each of the 3D points for every sub-block
to drive the iterative solution.

In \citep{zhang:17} the convergence of the consensus is analyzed for
the case where the points are unique in the sub-blocks and there can
be separate estimates for the cameras if they picture 3D points in
more than one sub-block. There are two findings: The first is not
relevant for our case as it pertains to the cameras, particularly
their rotation. The second is concerned with the translation of the
camera and the 3D points which is relevant also for our case. Here,
the condition for convergence consists in that the 3D points should
not be extremely close to the cameras, which can be safely assumed to
hold for SfM problems \citep{zhang:17}, i.e., in our case.

\section{Extended Consensus-Based Bundle Adjustment}
\label{sec:ecba}

When looking more closely at (\ref{eq:admm7}) one can see that it is
actually very similar to bundle adjustment using stochastic (ground)
control points for absolute pose estimation / orientation. In our
case, the role of the control points is taken by the TPs which link
two or more sub-blocks and we extend $f$ by weighting in the form of
the inverse of the approximate covariance matrix of the measurements
$\mathbf{C}^{-1}_{\mathbf{x}}$ as in (\ref{eq:normalw})

\begin{equation}
  \arg \min_{\mathbf{P}, \mathbf{X}} \Bigl( f(\mathbf{P},\mathbf{X},\mathbf{x},\mathbf{C}^{-1}_{\mathbf{x}}) + \| \mathbf{X} - \mathbf{X}_{\mbox{\scriptsize TP}} \|_{\mathbf{W}_{\mbox{\tiny TP}}} \Bigr) \; .
  \label{eq:batp}
\end{equation}

Opposed to (\ref{eq:admm7}), here the relative weighting between
$f(\mathbf{P},\mathbf{X},\mathbf{x},\mathbf{C}^{-1}_{\mathbf{x}})$ and
the distance between the 3D points $\mathbf{X}_j$ and the TPs is done
by a weight matrix $\mathbf{W}_{\mbox{\tiny TP}}$. The latter can be
used to introduce the information on the global 3D points common to
different sub-blocks, i.e., TPs, in a more detailed way.  We again
note that we split up the cameras unambiguously, i.e., every camera is
just contained in one sub-block, while the 3D points can be visible in
more than one sub-block. Written in the way of (\ref{eq:admm7}) with
$\mathbf{C}^{-1}_{(X_{\mbox{\tiny TP}})^t}$ the inverse of the
covariance information for a 3D point it reads

\begin{eqnarray}
  (\mathbf{P})^{t+1} & = & \arg \min_{\mathbf{P}, \mathbf{X}^l} \Bigl( f_l((\mathbf{P})^t,(\mathbf{X}^l)^t,\mathbf{x},\mathbf{C}^{-1}_{\mathbf{x}})
  + \| (\mathbf{X}^l)^t - (\mathbf{X}_{\mbox{\scriptsize TP}})^t \|_{\mathbf{C}^{-1}_{(\mathbf{X}_{\mbox{\tiny TP}})^t}} \Bigr) \; .
  \label{eq:bact}
\end{eqnarray}

When experimenting with the above ADMM scheme, we found similar to
\citet{ayeni:80} that a relative large number of iterations is needed
and we thought about ways to improve convergence.

Bringing together both above issues led to the idea to replace the
averaging of the corresponding 3D points in different sub-blocks in
(\ref{eq:admm8}) by adjustment / intersection / triangulation just for the
TPs: I.e., we take the cameras from the different sub-blocks of
(\ref{eq:bact}) as fixed and just optimize the position of the TPs
obtaining by this means also their covariance information
$\mathbf{C}_{(\mathbf{X}_{\mbox{\tiny TP}_j})^t}$ needed for
(\ref{eq:bact})

\begin{eqnarray}
  (\mathbf{X}_{\mbox{\scriptsize TP}_j})^{t+1}, \mathbf{C}_{(\mathbf{X}_{\mbox{\tiny TP}_j})^{t+1}} & = & \arg \min_{\mathbf{X}_j}
  f_j((\mathbf{P})^{t+1},(\mathbf{X}_j)^t,\mathbf{x}_k,\mathbf{C}^{-1}_{\mathbf{x}_k}) \; .
  \label{eq:bars}
\end{eqnarray}

Overall, we just switch between (\ref{eq:bact}) and (\ref{eq:bars})
without the need to compute an improvement as in
(\ref{eq:admm9}). This has two basic advantages:

\begin{enumerate}
  \item We found that the estimated $(\mathbf{X}_{\mbox{\scriptsize TP}_j})^{t+1}$ from 
    (\ref{eq:bars}) lead to a much better convergence than the averages
    from (\ref{eq:admm8}).
  \item One obtains an estimate of the covariance matrix
    $\mathbf{C}_{(\mathbf{X}_{\mbox{\tiny TP}_j})^{t+1}}$ which can be
    readily plugged into (\ref{eq:batp}) without any fiddling with the
    penalty parameter $\rho$, as the covariance matrix is inherently
    scaled correctly.
\end{enumerate}

Intuitively, the proof in \citep{zhang:17} should also apply here as
it is a refined version of the above ADMM scheme. Practically, we
found a very good convergence behavior of the above scheme as
presented in detail below in Section \ref{sec:ep}.

Additionally, the above scheme is related to resection / intersection schemes
such as \citep{bleser:07,mitra:08,pritt:14,hansch:16}, where one switches
between optimizing individual points and cameras. This can also be regarded as
Gauss-Seidel iteration \citep{ayeni:80,wang:01}.

For parallel processing, the above schemes have to distribute all
points and all cameras for each iteration.  Opposed to this, for the
computation of (\ref{eq:bars}) we just have to distribute all cameras
and the TPs. Additionally, as demonstrated below in the experiments by
extending the number of sub-blocks and thereby reducing the number of
cameras in each sub-block, the convergence rate is much better for
larger sub-blocks. Particularly, we found that one should adapt the
number of sub-blocks to the number of threads available on a computer.

Finally, we note the link to \cite{jeong:12}, which employ ``embedded
point iterations'' to speed up the convergence of their method for
bundle adjustment. As we similarly optimize the point positions
independently, we can also expect a corresponding speed-up.

\subsection{Refined Re-Weighting}
\label{sec:refrw}

The covariance for all cameras is computed by means of
(\ref{eq:bars}). As we use $\mathbf{C}_{(\mathbf{X}_{\mbox{\tiny TP}_j})^{t+1}}$ for weighting
the TPs in the sub-blocks, it seems reasonable to compute this per
sub-block. Particularly, the idea is to use per sub-block only the
information from all cameras in all sub-blocks besides the current,
i.e., $L\setminus l$.

While we formally work with the inverse of the covariance matrix, we
note that the covariance matrix is itself the inverse of the normal
equations which are derived from the Jacobian $\mathbf{A}$ using as
weight the inverse of the approximate covariance matrix of the
measurements for the 3D point $j$ $\mathbf{C}_j^{-1}$,
cf.~(\ref{eq:normalw}). Thus, we skip the double inverse and use as
refined weight matrix the normal equations reduced to those cameras
which are not included in sub-block $l$

\begin{eqnarray}
  \mathbf{C}^{-1}_{(\mathbf{X}^l_{\mbox{\tiny TP}_j})^{t+1}} & = & (\mathbf{A}_j^{L\setminus l})^{\mbox{\sf\scriptsize T}}
  (\mathbf{C}_j^{L\setminus l})^{-1}(\mathbf{A}_j^{L\setminus l})
  \; .
\end{eqnarray}

The above just applies to the calculation of the weight matrix. The
computation of the point position $(\mathbf{X}_{\mbox{\scriptsize
    TP}_j})^{t+1}$ is based on all cameras in all sub-blocks in which
the point is visible.

\subsection{Sub-Block Construction by Means of Graph Partitioning}
\label{sec:graphpartitioning}

Our parallelization builds on sub-blocks. Basically, for larger blocks
one usually employs as many sub-blocks as threads available on a
computer. Yet, below a certain block size it is more efficient to
adjust the block serially without parallelization.

Concerning the sub-blocks it is important to obtain a balanced load of
the threads particularly concerning computation time. Additionally, it
is helpful when the sub-blocks do not have too many TPs to reduce the
computation in (\ref{eq:bars}) and the effort for the exchange of the
information concerning the 3D points and cameras between
(\ref{eq:bact}) and (\ref{eq:bars}).

Particularly, we have defined this as a graph partitioning problem of
the visibility graph and similarly as \cite{ni:07} we use METIS'
\citep{karypis:98} 5.1 multilevel recursive bisection graph
partitioning algorithm to solve it.

Configuring METIS proved difficult and could only be done
empirically. Basically, the computational complexity for bundle
adjustment of a sub-block is related to the third power of the number
of cameras and the number of observations in the sub-block. The first
issue is that METIS does not allow for a constraint on the number of
elements in a sub-block. Additionally, because we use CG for solving
and as in most larger sub-blocks far from all cameras are directly
related by observations, the computational complexity is well below
the third power of the number of cameras.

We empirically found that weighting each node of METIS, i.e., camera,
by the cubic root of the sum of the number of observations of all
cameras for all 3D points a camera ``sees'' gives the best result in
terms of computation time. Intuitively, by this means we implicitly
model that the computational complexity for a camera is not only
related to the number of 3D points it has imaged, but also to how many
cameras it is related to, implicitly given by the number of
observations for the 3D point.

\subsection{Robust Estimation}
\label{sec:robest}

In numerous experiments we have found that a very simple approach works very
well for robust estimation given the results of a SfM pipeline such as
\citep{mayer:12} based on Random Sample Consensus -- RANSAC
\citep{fischler:81a}. The approach is based on normalized weighted squared
residuals $\overline{v_k}^2 = \frac{\mathbf{v}_k^{\mbox{\sf\scriptsize T}}
  \mathbf{C}_{\mathbf{x}_k} \mathbf{v}_k}{\sigma_0^2}$, i.e., the weighted
squared residual divided by the variance of the weighted image measurements
\begin{equation}
\sigma_0^2 = \frac{\sum_k{\mathbf{v}_k^{\mbox{\sf\scriptsize T}} \mathbf{C}_{\mathbf{x}_k}
  \mathbf{v}_k}}{redundancy} \; ,
  \label{eq:sigma0}
\end{equation}
and works in two steps:
\begin{itemize}
  \item After basic convergence of the adjustment, the observations
    for which $\overline{v_k}$ is beyond a certain threshold $t_v$
    (empirically set to 3) are weighted down strongly by 1.e-4.
  \item After finishing the iteration, the corresponding observations
    are deleted. If this means that a 3D point is observed in only one
    camera, it is completely eliminated.
\end{itemize}

We had initially \citep{mayer:12} used the variance of individual
weighted squared residuals $\sigma_{\mathbf{v}_k}^2$ instead of
$\sigma_0^2$, but found that this often lacks robustness. The latter
is also the case if one estimates $\sigma_0$ according to
(\ref{eq:sigma0}). We, thus, have switched to estimating $\sigma_0$
for robust estimation by means of the Median Absolute Deviation (MAD)
multiplied by the factor 1.4826 under the assumption that the variance
is normally distributed. Experimentally this gives rather robust
results, particularly when the estimation is done per camera, which is
not only faster but also more meaningful especially when the cameras
are of a different type, possibly with a rather different image size.

While it would be conceptually useful to employ only one consistent
robust estimation scheme, different schemes are used in the serial and
the parallel case. This basically accounts for the fact that depending
on the block structure a substantial number of 3D points might be TPs
and visible in a larger number of sub-blocks. When a point is seen in
a sub-block by only very few cameras, a reliable detection of an
outlier is difficult.

Thus, we decided to conduct robust estimation for the parallel case by
robust weighting during the estimation of the 3D points in the
intersection part, because there all cameras a point is seen in are
involved. Additionally, complete 3D points are eliminated instead of
just individual image observations. As shown later in the experiments
in Section \ref{sec:robust}, the difference between the effectiveness
of the parallel and the serial version is relatively small,
particularly when $t_v$ is set to 4 in the parallel case instead of 3
in the serial case as found by numerous experiments.

\subsection{Implementation Details}
\label{sec:implement}

Our implementation is influenced in many ways by \cite{jeong:12}. We
use the Schur complement to obtain the RCS which is solved by
preconditioned CG. By employing outer products we accumulate the 3D
points directly in the RCS and avoid intermediate matrices. We also
follow \cite{jeong:12} by utilizing block-based Jacobi
preconditioning, i.e., we use the inverses of the diagonal blocks for
preconditioning. Similarly as \cite{jeong:12} we employ a three
parameter Rodriguez representation but limited to the incremental
rotation during adjustment, avoiding problems with rotations close to
180$^{\circ}$.  For the LM algorithm we also multiply the diagonal of
the normal equations $\mathbf{N}$ (\ref{eq:normal}) with the damping
factor, i.e., $\mathbf{N}_{aug} = \mathbf{N} + \lambda \mathbf{I}\;
\mbox{diag}\{\mathbf{N}\}$.

When developing and later improving \citep{mayer:12} we found in many
experiments that setting $\lambda$ to 1.e-3 or 1.e-4 is sufficient (in
this paper it is set to 1.e-4, but for wide-baseline settings it is
advantageous to set it to 1.e-3). Only for finally estimating
covariance information---if needed---we set $\lambda$ to 1.e-6 after
initial convergence. For the determination of convergence we compare
the best reprojection error obtained so far with the current
reprojection error. If the ratio is below 1.01, we stop the
iteration. To deal with convergence problems particularly in the first
round of iterations, we allow for one iteration without an improvement
of more than 1.01.

We acknowledge that setting $\lambda$ to 1.e-4 leads to biased results
if covariance matrices are estimated. Yet, we note that we found this
relatively high value necessary to obtain convergence for difficult
datasets. A solution could be adapting $\lambda$ as in
\citep{jeong:12}, but that means another solution of the normal
equations with the corresponding effort. Revisiting this issue could
be part of future work although our current experiences do not point
to significant deficits in this respect.

We obtain basic robustness concerning far-away points as well as
outliers by throwing out the very few points where the intersection of
a 3D point does not converge.

We finally note that our implementation which is based on METIS
\citep{karypis:98} and the linear algebra library Eigen is available on
GitHub: \url{https://github.com/helmayer/RPBA}

\section{Experiments}
\label{sec:ep}

As a baseline we compare our (extended) approach to PBA -- Parallel
Bundle Adjustment \citep{wu:11}, but the main focus is on the
comparison to the (plain) consensus-based approach introduced in
Section \ref{sec:cba}.  For the experiments we use a computer with 12
cores, 24 threads and 64 GB of memory running Linux.

We use the data sets Final 13682 and Final 961 introduced in
\citep{agarwal:10} and Dino from \citep{seitz:06} as well as a
Simulated data set.  The latter is in the form of a ``traditional''
aerial block with 50 strips and 400 cameras per strip, 0.6 endlap, 0.2
sidelap and around 100 statistically distributed points per camera but
only in the overlap area, i.e., 20,000 cameras and 1,861,000 3D
points. We added 1.0 pixels Gaussian noise to the image observations
and perturb the camera parameters by 0.0001 radians in the angles and
$0.1$ units in the coordinates, leading to an initial average back
projection error of about 30 pixels.

\begin{table}
  \begin{tabular}{lllll}
    data set & \# cameras & \# points & \# observations & \# parameters \\
             &           &           &                 & for cameras \\
    \noalign{\smallskip}
    \hline
    \noalign{\smallskip}
    Final 13682  & 13,682          &  4,456,117 &  57,973,736 & n * (6 + 2) \\
    Final 961    & 961             &    187,103 &   3,385,950 & n * (6 + 2) \\
    Dino         & 363             &     37,796 &     423,718 & n * 6 + 7 \\
    Simulated    & 50 $\times$ 400 &  1,861,000 &  12,001,562 & n * 6 \\
  \end{tabular}
   \caption[]{Characteristics of the data sets}
   \label{tab:datasets}
\end{table}

For the two Final data sets we estimate the camera constant /
principal distance / focal length as well as the first (quadratic)
parameter of the radial distortion together with the camera poses
separately for each camera as has been done in other experiments with
these data sets and is also standard with PBA. For the Simulated data
set we just compute the camera poses, but for Dino we estimate the
camera poses as well as a joint calibration matrix and quadratic as
well as quartic radial distortion.

As due to reasons of efficiency it is not meaningful to work with too
small sub-blocks, we limited their minimum size for the experiments to
an empirically found value of 70 cameras.

Our approach used in the experiments is the extended consensus-based
adjustment approach introduced in the preceding section including the refined
re-weighting of Section \ref{sec:refrw}. It is called the ``extended
approach'' for the remainder of this paper. Compared to this, we call the
consensus-based approach from Section \ref{sec:cba} the ``plain approach''.
An overview of all employed approaches is given in Table \ref{tab:approaches}.

\begin{table}
  \begin{tabular}{ll}
    extended approach & Section \ref{sec:ecba} incl.~refined re-weighting (Section \ref{sec:refrw}) \\
    plain approach & Section \ref{sec:cba} \\
    PBA & \citep{wu:11}
  \end{tabular}
   \caption[]{Overview of the approaches}
   \label{tab:approaches}
\end{table}

\subsection{Sub-Block Sizes and Computation Times}
\label{sec:split}

In this section we give a visual impression of the data sets as well
as how they are split into sub-blocks by METIS \citep{karypis:98} via
the graph partitioning approach introduced in Section
\ref{sec:graphpartitioning}.

\begin{figure}
   \begin{center}
      \leavevmode \unitlength 1cm 
      \includegraphics[width=11cm]{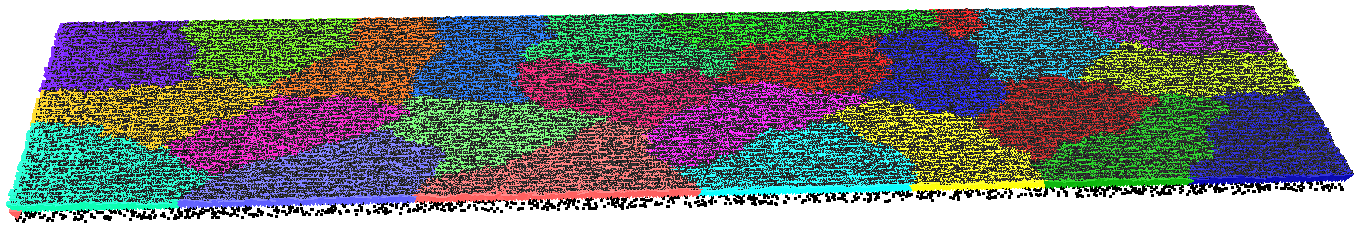}
      \vspace{0.3cm}
      \includegraphics[width=12cm]{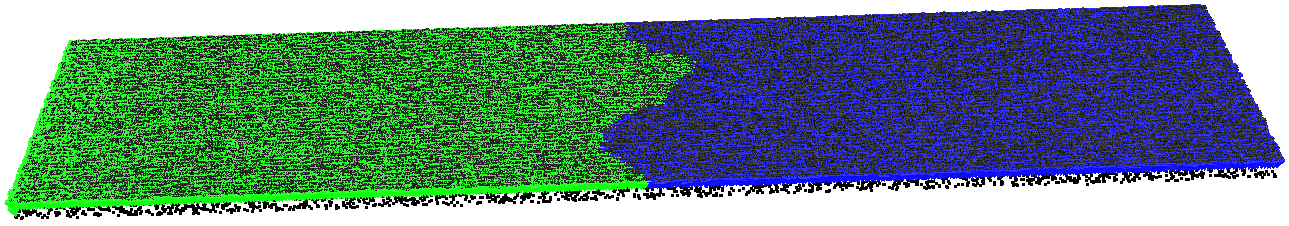}
   \end{center}
   \caption[]{Simulated block with 50 strips and 400 cameras per strip
     split by METIS into 24 sub-blocks (top) and 2 sub-blocks (bottom),
     marked by different colors.}
   \label{fig:simulation}
\end{figure}

\begin{figure}
   \begin{center}
      \leavevmode \unitlength 1cm 
      \includegraphics[width=7cm]{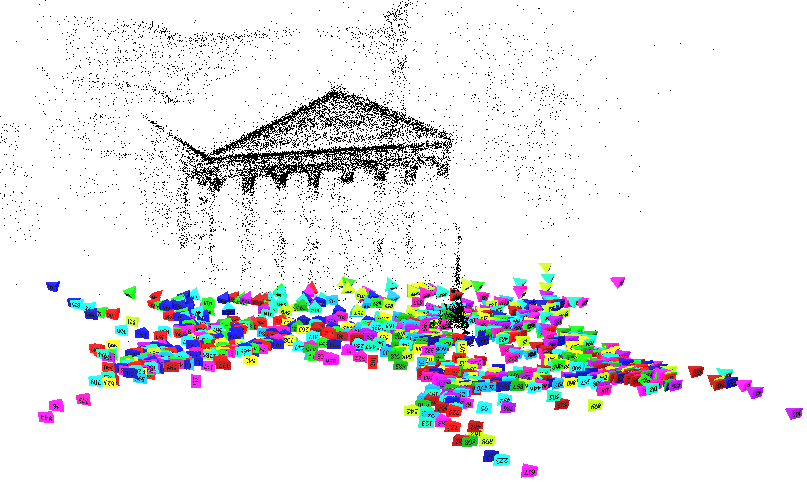}
   \end{center}
   \caption[]{Final 961 from \citep{agarwal:10} split by METIS into 14
     sub-blocks marked by different colors.}
   \label{fig:final961}
\end{figure}

\begin{figure}
   \begin{center}
      \leavevmode \unitlength 1cm 
      \includegraphics[width=12cm]{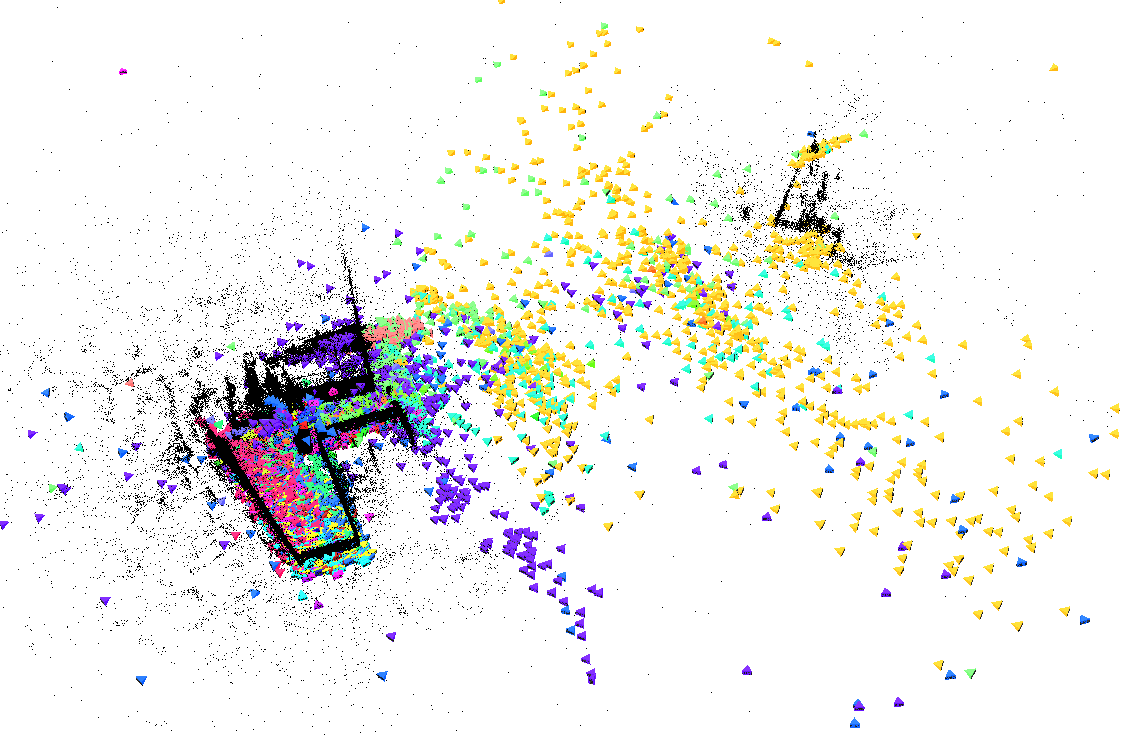}
   \end{center}
   \caption[]{Final 13682 from \citep{agarwal:10} split by METIS into 24
     sub-blocks marked by different colors.}
   \label{fig:final13682}
\end{figure}

\begin{figure}
   \begin{center}
      \leavevmode \unitlength 1cm 
      \includegraphics[width=6cm]{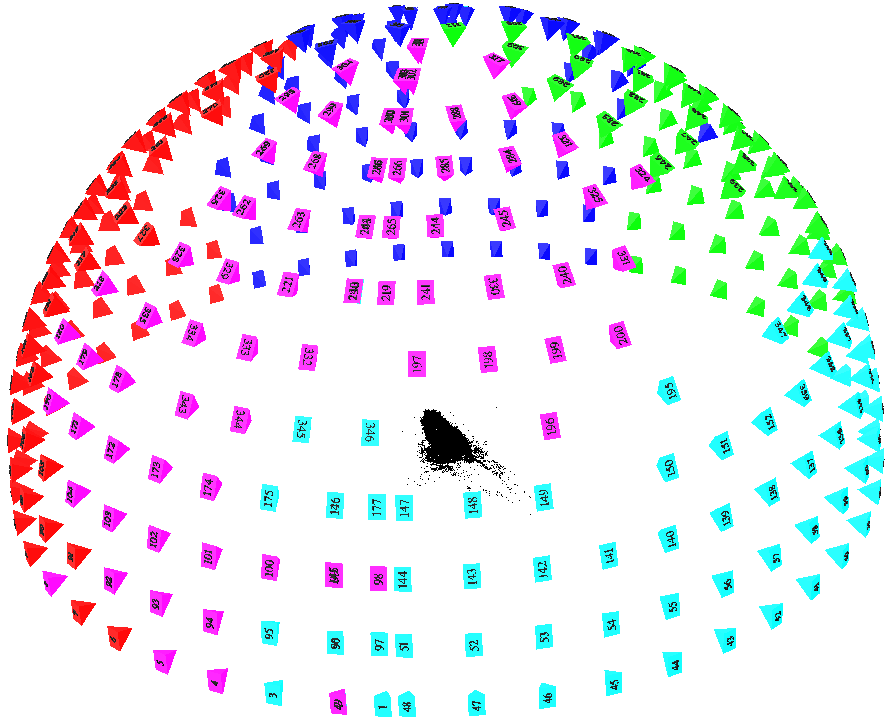}
   \end{center}
   \caption[]{Data set Dino from \citep{seitz:06} split by METIS into
     5 sub-blocks marked by different colors.}
   \label{fig:dino}
\end{figure}

Figure \ref{fig:simulation} shows how the Simulated data set with 50 strips is
split into 2 and 24 sub-blocks. In the second case, this corresponds to the
number of physical threads of our processor. One can see that the algorithm
splits the data set into compact, mostly contiguous sub-blocks. The latter is
not so obvious for the two Final data sets (cf.~Figures \ref{fig:final961} and
\ref{fig:final13682}), because the images are strongly overlapping as most try
to picture the same part of the scene although from different view
points. While for Final 961 nearly no structure can be seen, due to its much
larger size, Final 13682 shows some more contiguous areas (e.g., the sub-block
marked in yellow). Finally, Figure \ref{fig:dino} gives an impression of the
data set Dino, which is used to demonstrate the efficiency of our approach for
robust estimation in Section \ref{sec:robust}. As it consists of merely 363
cameras, it is split into only 5 sub-blocks due to the minimum sub-block size
of 70. Also here many cameras see the same area of the scene (there are points
projected into more than 90 cameras), but still the sub-blocks are mostly
contiguous.

\begin{figure}
   \begin{center}
      \leavevmode \unitlength 1cm 
      \includegraphics[width=10cm]{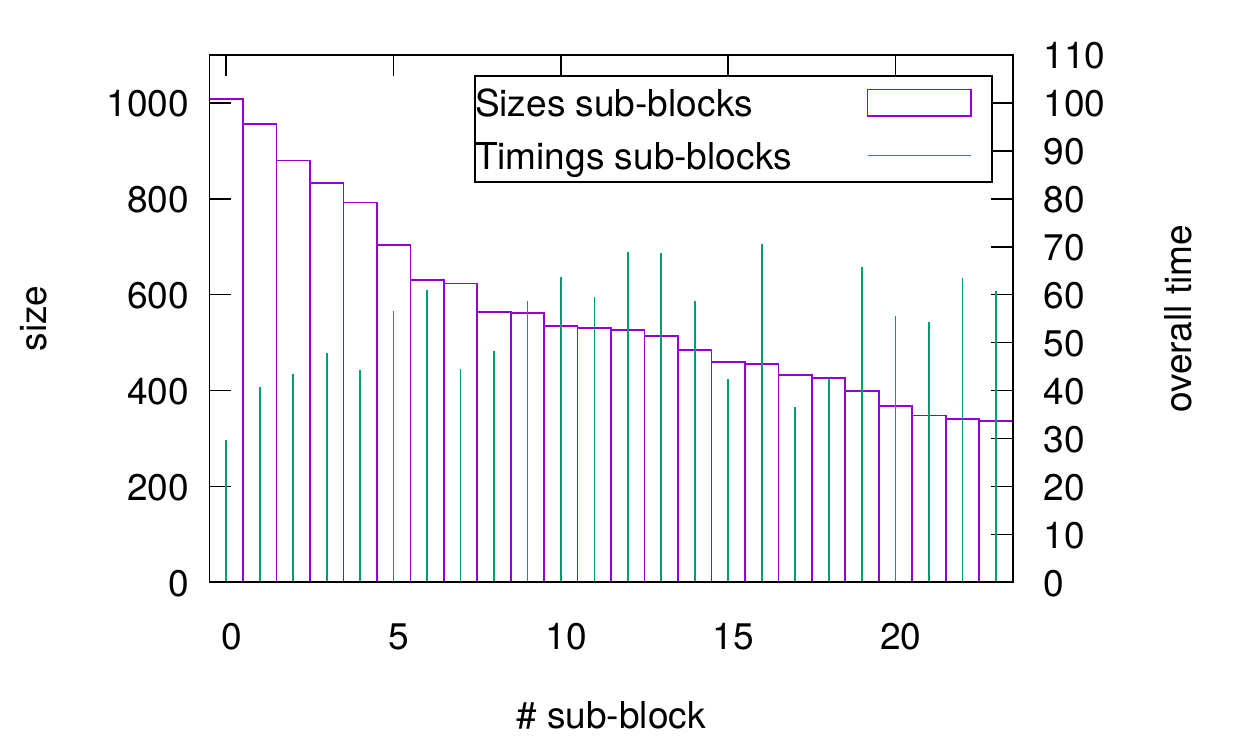}
   \end{center}
   \caption[]{Sizes of the sub-blocks (box) ordered according to size
     in comparison with the overall time per sub-block/thread
     (vertical line) for data set Final 13682.}
   \label{fig:subblocks}
\end{figure}

In Figure \ref{fig:subblocks} we give a comparison between the sizes of the
sub-blocks and the overall computation times for all iterations ordered
according to the sub-block sizes for Final 13682. One can see that there is no
direct relation between the two: larger blocks might have a smaller
computation time than much smaller blocks. We assume that this is due to the
different neighborhood structure and, thus, sparsity of the
individual sub-blocks. Overall, this is the optimum result concerning the computation
time we found by numerous experiments particularly for this large data set. As
outlined in Section \ref{sec:graphpartitioning}, the basic reason why we could
not do better has been that for the graph-based METIS we could not define
constraints on the overall number of cameras and observations per
sub-block. We found similar distributions also for numerous other data sets.

\subsection{Convergence Behavior Depending on the Number of Threads}
\label{sec:convergence}

This section gives a comparison of the convergence behavior and
corresponding timings for different numbers of threads for the
extended approach based on the data set Final 13682. Convergence is
described according to the behavior of the average standard deviation
of the weighted image measurements $\sigma_0$ (\ref{eq:sigma0}) which
can be interpreted as the average error of the weighted image
measurements. The latter holds because we use normalized weights which
are 1 on average. This is trivial for the two Final datasets and the
Simulated dataset were no weights are given and they are, thus, simply
set to 1. For Dino we determine individual weights by means of least
squares matching and then normalize them.

\begin{figure}
   \begin{center}
      \leavevmode \unitlength 1cm 
      \includegraphics[width=10cm]{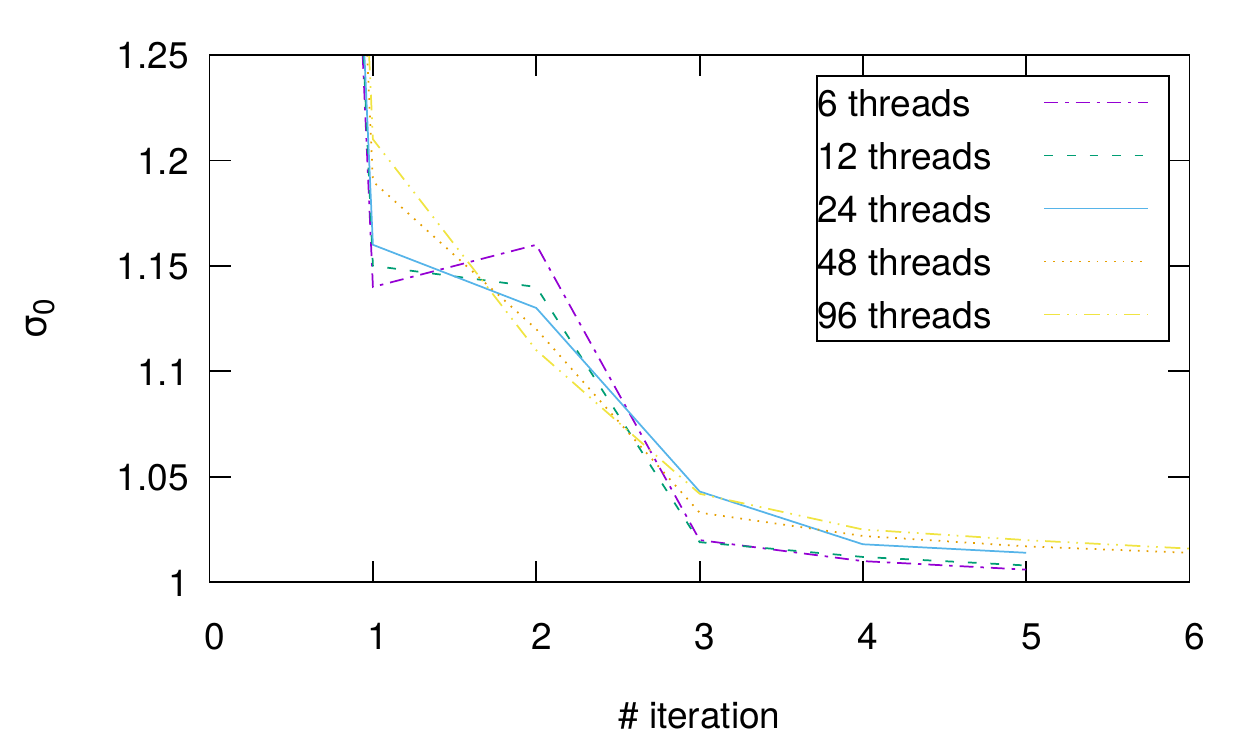}
   \end{center}
   \caption[]{Convergence of Final 13682 for different numbers of
     threads.}
   \label{fig:threads-convergence}
\end{figure}

Figure \ref{fig:threads-convergence} shows that there is not necessarily
convergence for all iterations, an iteration consisting of one solution of
equations (\ref{eq:bact}) and (\ref{eq:bars}) each. Particularly in the first
couple of iterations it is possible that coordinates for joint 3D points in
different sub-blocks can be quite different leading to large errors. Yet, as
shown in Table \ref{tab:13682}, in the end the differences between the various
solutions are very small in terms of $\sigma_0$.

Overall, the convergence is faster for a smaller number of sub-blocks,
which is to be expected as there are less common 3D points which have
to be adjusted by means of the interaction between bundle adjustment
for the sub-blocks and intersection of the 3D points. On the other
hand, the convergence is more smooth, particularly without any rise,
for a larger number of sub-blocks. Finally, please note that the
fastest solution is obtained for the physical number of threads on the
employed computer, namely 24. While for a smaller number not all the
computing power is used, a larger number implies more not really
efficient iterations with bundle adjustment of sub-blocks and
intersection of the 3D points. It is, thus, recommended to split the
block in as many sub-blocks as physical threads are available as long
as the sub-blocks do not become too small and, thus, the overhead for
parallel computing becomes too large.

\begin{table}
  \begin{tabular}{lllll}
    threads & \# iterations & $\sigma_{0}$ [pixel] & time [sec] & convergence $\sigma_{0}$ \\
    \noalign{\smallskip}
    \hline
    \noalign{\smallskip}
    6      & 6  & 1.006 & 473 & 2.91 1.14 1.16 1.020 1.010 \\
    12     & 6  & 1.008 & 219 & 2.91 1.15 1.14 1.019 1.012 \\
    24     & 6  & 1.014 & \bf 149 & 2.91 1.16 1.13 1.043 1.018 \\
    48     & 7  & 1.014 & 212 & 2.91 1.19 1.12 1.033 1.022 1.017 \\
    96     & 7  & 1.016 & 392 & 2.91 1.21 1.11 1.042 1.025 1.020 \\
  \end{tabular}
   \caption[]{Convergence of Final 13682 from \citep{agarwal:10} for different
     numbers of threads. There are 24 threads available by the hardware.}
   \label{tab:13682}
\end{table}

\subsection{Comparison to PBA and Between Different Variants of Consensus}
\label{sec:compare}

In this section we compare our extended approach (cf.~Table
\ref{tab:approaches}) with the baseline PBA \citep{wu:11} in its
current version (master-PBA --
\url{http://sourceforge.net/PBA-master}) as well as our implementation
of the plain approach based on the standard deviation / average error
of the weighted image measurements $\sigma_0$ (cf.~preceding section).

Concerning PBA it was not easy to process the Final 13682 data
set. Only by switching to ``measurement distortion'' instead of
``projection distortion'' we could obtain a solution which was not
extremely much worse than for the other two approaches. On the other
hand, the solution by PBA for Final 961 is very close to the solution
obtained by our extended consensus. The same is true for the Simulated
data set. We let PBA write its solution with the improved parameters
into a file and used our software to consistently compute $\sigma_0$.

For the plain approach we had to fix a couple of parameters.  Of
biggest importance we found to be the parameter $\rho$ (cf.~Section
\ref{sec:cba} equation (\ref{eq:admm4})). As recommended in
\citep{zhang:17}, we modify $\rho$ by multiplying it with a factor
$1.01$ for each iteration. The latter is the same for all data
sets. For $\rho$ itself different values were necessary for different
data sets. For Final 13682 and 961 we set $\rho$ to 1000, for the
Simulated data set to 200. We additionally found that the plain
approach could not deal with a few very far-away as well as with
outlier points. To deal with this, we included a filter throwing out
points with very large coordinate values ($>$ 1.e10) as well as a
large distance between projected and measured points. Yet, this only
applied to a few points.

\begin{figure}
   \begin{center}
      \leavevmode \unitlength 1cm 
      \includegraphics[width=10cm]{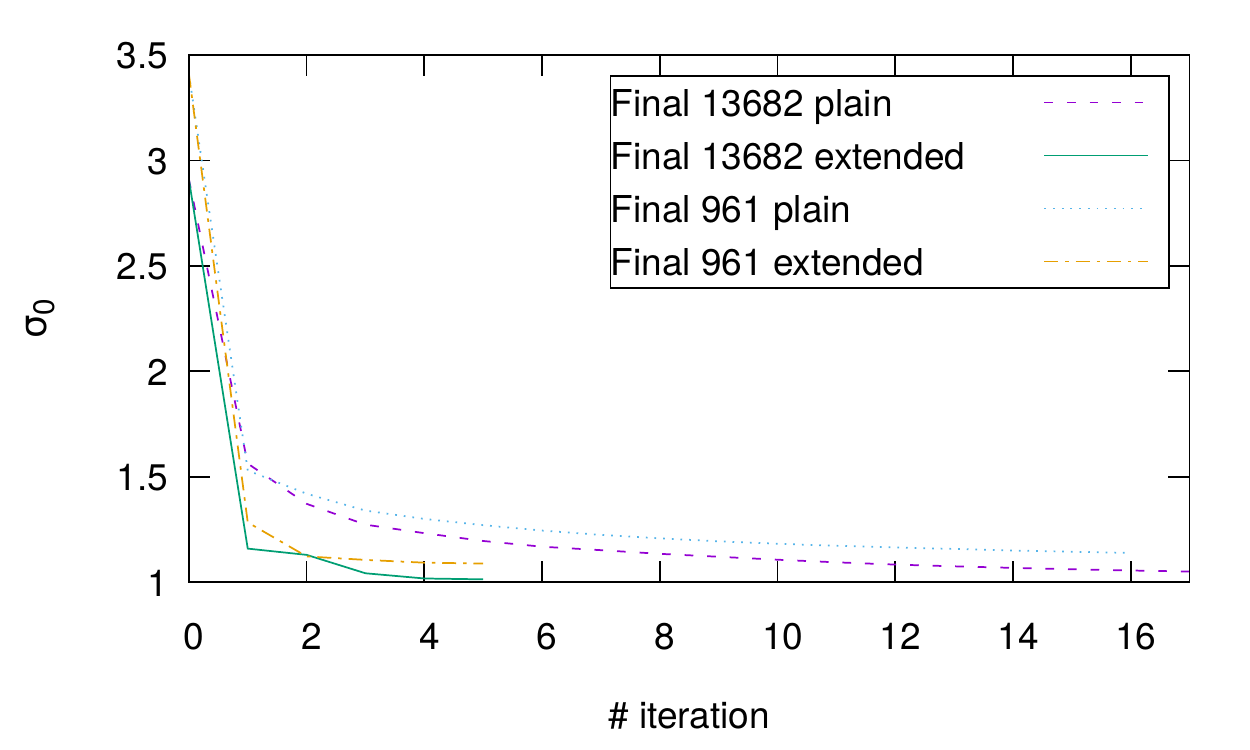}
   \end{center}
   \caption[]{Convergence Final 13682 and 961 for plain and extended
     approach (all on 24 threads).}
   \label{fig:consense-convergence}
\end{figure}

Table \ref{tab:consensus} summarizes the results for the data sets
Final 13682 and 961. One can see that our extended approach is fastest
and produces the best results for Final 13682. The result is very
similar to the result of PBA for Final 961. As discussed above, we
could not find any setting for the parameters for PBA which gave a
better result for Final 13682. Concerning the comparison of the plain
and the extended approach, the latter is not only faster, but it
produces also results of a higher accuracy in much fewer
iterations. Figure \ref{fig:consense-convergence} gives a graphical
impression of the convergence behavior. It shows that due to the use
of the additional information in the form of the covariance matrix of
the TPs as well as due to the replacement of the simple averaging by
triangulation based on adjustment, the extended approach converges
considerably faster and reaches a lower value for $\sigma_0$.

\begin{table}
  \begin{tabular}{lllll}
    data set & approach & \# iterations & $\sigma_{0}$ [pixel] & time [sec] \\
    \noalign{\smallskip}
    \hline
    \noalign{\smallskip}
    13682  & plain    & 18  & 1.051 & 443 \\
           & extended & 6  & 1.014 & \bf 149  \\
           & PBA      &    & 1.123 & 175  \\
    \noalign{\smallskip}
    \hline
    \noalign{\smallskip}
    961    & plain    & 17  & 1.139 & 13.3 \\
           & extended & 6   & 1.089 & \bf 7.4  \\
           & PBA      &     & 1.088 & 18.2
  \end{tabular}
   \caption[]{Comparison of the plain and the extended approach as well as PBA
     \citep{wu:11} for Final 13682 and 961.}
   \label{tab:consensus}
\end{table}

For the Simulated data set all approaches work relatively well
(cf.~Table \ref{tab:simulation}).  We first note that the differences
of the obtained $\sigma_{0}$ can basically be neglected.  Both, the
plain approach and PBA are considerably slower than the extended
approach.  We have also compared the results for the extended approach
for different numbers of threads. As above for Final 13682, one can
see that we achieve the optimum speed for the physical number of
threads (24) of the computer used for the experiments.

\begin{table}
  \begin{tabular}{llll}
    approach / threads & \# iterations & $\sigma_{0}$ [pixel] & time [sec] \\
    \noalign{\smallskip}
    \hline
    \noalign{\smallskip}
    plain 24 & 8  & 1.0008 & 27.2 \\
    \noalign{\smallskip}
    \hline
    \noalign{\smallskip}
    extended 1      &    & 0.9978 & 210  \\
    extended 2      & 4  & 1.0005 & 36.7 \\
    extended 3      & 4  & 1.0005 & 27.3 \\
    extended 6      & 4  & 1.0006 & 16.9 \\
    extended 12     & 4  & 1.0007 & 12.9 \\
    extended 24     & 4  & 1.0008 & {\bf 12.4} \\
    extended 48     & 4  & 1.0008 & 16.3 \\
    extended 96     & 4  & 1.0010 & 29.1  \\
    \noalign{\smallskip}
    \hline
    \noalign{\smallskip}
    PBA 24    &    & 1.0002 & 36.4
  \end{tabular}
   \caption[]{Comparison of plain and extended approach as well as PBA
     for the Simulated data set. For the extended approach (center
     rows) the results for different numbers of threads are given. 24
     is the number of physical threads.}
   \label{tab:simulation}
\end{table}

\subsection{Ablation Study}
\label{sec:ablation}

This section quantifies the influence of our contributions based on
the Simulated data set as well as Final 961 and 13682.  In Table
\ref{tab:ablation} we compare the result for the complete extended
approach with simplified versions as well as with one version of the
plain approach improved by means of the refined re-weighting of
Section \ref{sec:refrw}, but without the intersection of the 3D
points.

To quantify the influence of the refined re-weighting introduced in
Section \ref{sec:refrw} and included in the complete system, we
compare it by weighting just by means of a scalar and by weighting
taking into account all cameras instead of just those in which a 3D
point in a certain sub-block is not visible as in Section
\ref{sec:refrw}. For weighting by means of a scalar, we use the same
scheme as for the plain approach, yet, we still compute the new point
estimates by means of intersection / adjustment of the 3D points. We
replace the full covariance matrix per point by a diagonal matrix
where all elements are the same and correspond to the $\rho$ of the
plain approach.

The full extended approach gives the best results concerning the
number of iterations and computation time. Weighting with all cameras
gives a little bit worse results. The accuracy is slightly lower, the
number of iterations as well as the time is a little bit higher. If
one substitutes the inverse covariance matrix by a scalar, the
accuracy particularly for Final 961 becomes substantially lower even
though the number of iterations and the computation time is similar to
the (full) extended approach. For Final 13682 the result is worse and
the number of iterations as well as the computation time are higher.

On the other hand, the results for the plain approach extended by the
refined re-weighting of Section \ref{sec:refrw} show that the plain
approach can be improved by the refined re-weighting. Yet, the result
for Final 13682 makes clear that the 3D intersection of the points is
essential and refined re-weighting is not enough. While the approach
converged very fast to a $\sigma_0$ of 1.04 pixels, it diverged after
this point and we did not find any means to fix this.

\begin{table}
  \begin{tabular}{llll}
    extended approach             & Simulated  & Final 961 & Final 13682  \\
    \# iterations                 &      4     &    6       &  6    \\
    $\sigma_{0}$                   & 1.0008      &  1.089   &  1.014  \\
    time [secs]                   &      12.4     &    7.4       &  149   \\
    \noalign{\smallskip}
    \hline
    \noalign{\smallskip}
    \multicolumn{4}{l}{extended: weighting with all cameras} \\ 
    \# iterations                 &      4      &    7       &  8    \\
    $\sigma_{0}$                   &  1.0008     &  1.101   &  1.020  \\
    time [secs]                   &      13.1    &    8.2       & 171    \\
    \noalign{\smallskip}
    \hline
    \noalign{\smallskip}
    \multicolumn{4}{l}{extended: weighting with scalar} \\ 
    \# iterations                 &      4      &    6       &  7    \\
    $\sigma_{0}$                   &  1.0009     &  1.157   &  1.044  \\
    time [secs]                   &      14.0    &    7.4       & 181    \\
    \noalign{\smallskip}
    \hline
    \noalign{\smallskip}
    \multicolumn{4}{l}{plain: refined re-weighting of Section \ref{sec:refrw}} \\ 
    \# iterations                 &      4      &    8       &  4 (diverging)   \\
    $\sigma_{0}$                   &  1.0009     &  1.124     &  1.055 \\
    time [secs]                   &      15.2    &    10.4    & (132)    \\
  \end{tabular}
   \caption[]{Results for different variants of the extended
     (original: top row) and the plain approach.}
   \label{tab:ablation}
\end{table}

\subsection{Effectiveness of Robust Estimation}
\label{sec:robust}

Concerning our approach for robust estimation introduced in Section
\ref{sec:robest}, we present two different types of experiments: The
first gives an impression of the improvement of accuracy possible by
means of robust estimation as well as what happens if there are
actually no outliers as for the Simulated data set. The second type of
experiments compares multiple use of our parallel robust (extended)
approach with standard serial bundle adjustment using just a single
thread. If the parallel approach would be significantly worse, the
multiple use should lead to substantially different final results.

For the former type of experiments, Table \ref{tab:robust} presents
the results. The improvements by means of robust estimation are
assessed by two means: First, based on the standard deviation /
average error of the weighted image measurements $\sigma_0$ and second
based on the variances of the parameters. For the latter, we give mean
and standard deviation as well as the median of the ratios of the
variances for the parameters estimated without and with robust
estimation.

For the Simulated data set for which we added a noise of 1.0 pixel a
very good approximation of 1.0006 pixels is obtained. Of about 12
Million observations less than 600 are eliminated, i.e., 0.005\%. This
verifies that our robust parallel approach does not (nearly) randomly
delete points.

The results for Final 961 and 13682 show that the accuracy could be
improved by a factor of 1.5 by eliminating less than 3\% of the
points. From other experiments we know that these are typical
values. The number of iterations and the computation time
approximately doubles for the robust case. For Final 961, the result
for the parallel approach is similar to that of the serial approach,
but the computation time is much lower.

The ratio of the variances match the ratio of the $\sigma_0$ for the
serial estimation result of Final 961, the only data set for which
variances could be calculated due to restrictions of the memory size
(64 GB were not enough). For parallel robust estimation, the ratios of
the variances fluctuate notably, but the average and particularly the
median of the ratios is similar to the ratio of the $\sigma_0$,
meaning that not only the average error but also the variances of the
parameters have improved (at least on average).

\begin{table}
  \begin{tabular}{llll}
    Parallel                       & Simulated  & Final 961 & Final 13682  \\
    \noalign{\smallskip}
    \hline
    \noalign{\smallskip}
    \# threads                      & 24            & 14               & 24 \\
    \# iterations                   & 5             & 11              & 10 \\
    $\sigma_{0}$ [pixel] not robust  & 1.0008        &  1.089          &  1.014  \\
    \# observations                 & 12,001,562    &  3,385,950      & 57,973,736 \\
    \# deleted (\%)                 & 576 (0.005\%) &  97,894 (2.9\%) &  1,503,098 (2.6\%) \\
    $\sigma_{0}$ [pixel] final       & 1.0006        & 0.721           &  0.706   \\
    Ratio $\sigma_0$ not robust / final &  0.9998    &  0.662          & 0.696 \\ 
    Ratio variances mean            &  0.9999 $\pm$ 0.005    &  0.793 $\pm$ 0.36 &  0.744 $\pm$ 0.27 \\ 
    Ratio variances median          &  0.9996       &  0.733 &  0.721 \\ 
    time [sec]                      & 16.7          & 14.9            & 266 \\
    \noalign{\smallskip}
    \hline
    \noalign{\smallskip}
    Serial \\
    \# deleted (\%)           & 0 (0.\%)    &  98,632 (2.9\%)   &  -- \\
    $\sigma_{0}$ [pixel] final & 1.0010      & 0.706            &  --   \\
    Ratio $\sigma_0$ not robust / final &  1.0002  & 0.648             & -- \\
    Ratio variances mean               &  --       & 0.627 $\pm$ 0.007 & -- \\ 
    Ratio variances median             &  --       & 0.630            & -- \\ 
    time [sec]               & 382          & 391              & --
  \end{tabular}
   \caption[]{Results for parallel robust estimation for different
     data sets. For Final 961 only 14 threads are used as the minimum
     number of cameras per sub-block/thread is set to 70. Because of
     lack of memory, no serial solution could be computed for Final
     13682 and no variance information for the serial solution for
     the Simulated dataset (--).}
   \label{tab:robust}
\end{table}

For efficiently testing the effectiveness of our approach for robust
estimation, we use the data set Dino and a version of our SfM software
which repeatedly adds small blocks to large blocks leading to many
bundle adjustments. Although we limited the minimum number of cameras
per sub-block / thread to 70 cameras, we still could use the parallel
bundle adjustment 31 times of overall 361 adjustments with between 2
and 5 threads (Dino consists of 363 cameras).

Table \ref{tab:dino-char} gives the results for serial robust bundle
adjustment as well as for the robust version of our extended approach.
We, particularly, compare the accuracy obtained before the final
(serial -- as benchmark) bundle adjustment, the accuracy after the
final bundle adjustment, the number of points obtained and the time it
takes to merge the block. One can see that the results apart from the
computation time are actually very similar. This is confirmed by Table
\ref{tab:dino-stat} which shows that the ratios of the variances of
the parameters for the parallel and the serial solution are close to
one and, thus, the variances are rather similar.

\begin{table}
  \begin{tabular}{lllll}
    approach & $\sigma_{0}$ [pixel] before & $\sigma_{0}$ [pixel] final & \# observations & time [sec] \\
             & final adjustment            &                            &                 & overall \\
    \noalign{\smallskip}
    \hline
    \noalign{\smallskip}
    serial    & 3.4  &  0.41  & 423,612 & 155.6 \\
    parallel  & 3.0  &  0.40  & 424,418 & 53.1
  \end{tabular}
   \caption[]{Comparison of the characteristics of the results for
     data set Dino processed by serial and parallel bundle adjustment,
     the latter with 2 to 5 threads in 31 of 361 adjustments (as the
     minimum number of cameras per thread is set to 70). The timings
     additionally contain a constant part for least-squares matching.}
   \label{tab:dino-char}
\end{table}

\begin{table}
  \begin{tabular}{llll}
    min & max & median & mean \\
    \noalign{\smallskip}
    \hline
    \noalign{\smallskip}
    0.91   & 1.16 &  0.981  & 0.983 $\pm$ 0.023
  \end{tabular}
   \caption[]{Statistics of ratios of variances of parallel and serial
     solution for Dino.}
   \label{tab:dino-stat}
\end{table}

\begin{figure}
   \begin{center}
      \leavevmode \unitlength 1cm 
      \includegraphics[width=12cm]{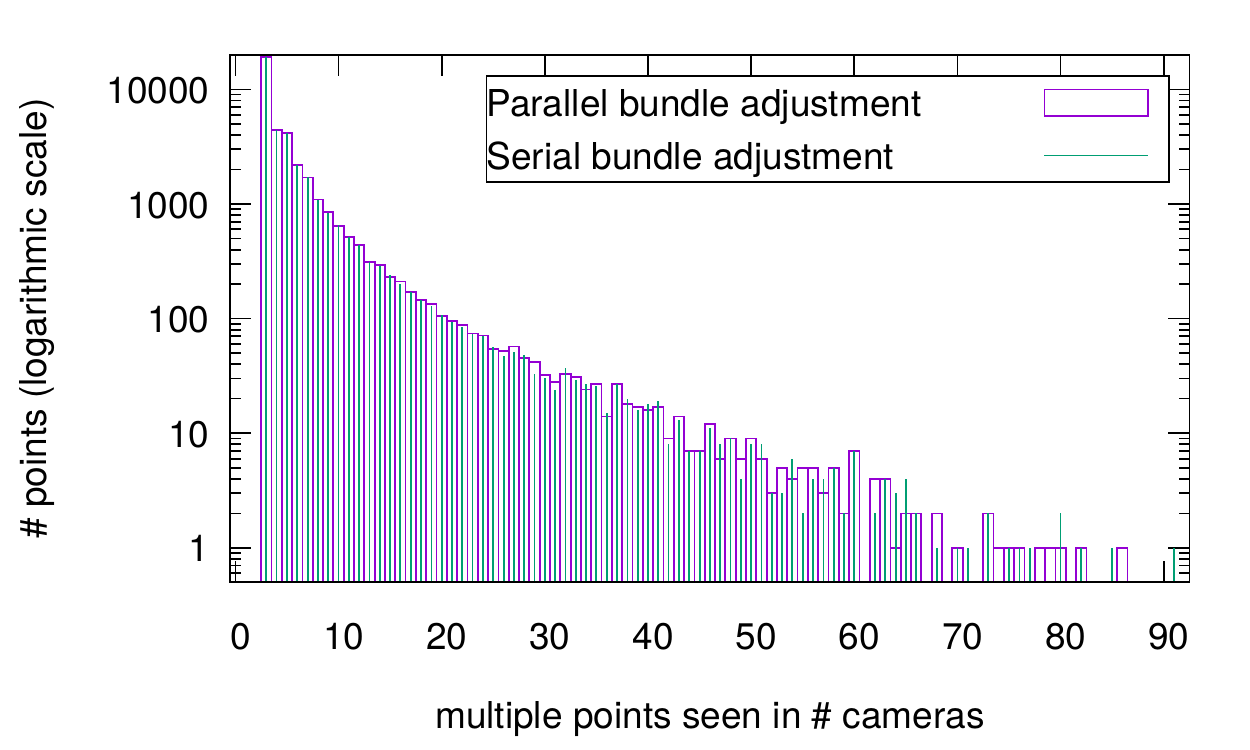}
   \end{center}
   \caption[]{Distribution of points projected in multiple cameras for
     data set Dino processed by serial and parallel robust bundle
     adjustment.}
   \label{fig:dino-distrib}
\end{figure}

To present the results a little bit more in detail, Figure
\ref{fig:dino-distrib} gives a comparison of the number of points
projected in a certain number of cameras. The more points are projected
in as many cameras as possible, the better the block stability becomes,
as they tie the cameras to each other.  The comparison shows that the
results of the serial and the parallel approach are very
similar. Together with the above comparison of the accuracies and
final variances, number of points and computation time this
demonstrates that the robust extended approach can be a fast
replacement for standard serial robust bundle adjustment.

\section{Conclusion}
\label{sec:cc}

Extending recent consensus-based approaches, we present means to
efficiently exploit the characteristics of modern parallel computers
for robust bundle adjustment. While resection / intersection schemes
have a weak convergence behavior and normal bundle adjustment needs
too many resources and is not inherently parallel, our scheme allows
to make use of each thread independently and still obtain a high
quality result in only a few iterations.

The latter distinguishes it from competing recent consensus-based
approaches, although there are just two big differences between their
work and ours:
\begin{itemize}
  \item They average the 3D point coordinates and use a penalty
    parameter $\rho$.
  \item We determine the points by forward intersection and get rid of
    the dataset-dependent penalty parameter $\rho$ by replacing it by
    the covariance information for the 3D points. Thus, no parameter
    tuning is necessary.
\end{itemize}
Interestingly, \cite{jeong:12} have also come to the conclusion that
adjusting 3D points can be particularly useful and use ``embedded
point iterations'' as an important part of their work.

Future work could comprise an improved rescaling of the dampener in
the Levenberg-Marquardt algorithm as recommended in \citep{jeong:12}
by means of trust region control \citep{lourakis:05}. Even though
splitting up the block in sub-blocks reduces efficiency issues while
solving the reduced camera system (RCS) considerably, for extremely
large (sub-)blocks still an ordering of the variables to minimize the
amount of fill-in during the solution could be helpful. For this,
exact minimum degree ordering as used in \citep{jeong:12} could
be an option.

Concerning robust estimation, it might be possible to just detected
the outliers during intersection and then forward this information to
be employed in resection. In addition to the implementation effort it
will still not be as statistically efficient as serial robust
estimation, because each sub-block contains at least for a certain
fraction of the 3D points only parts of the image observations.


\appendix

\section{Bibliography}

  \bibliography{parallel-bundle-19}

\end{document}